\documentclass[sigconf,screen]{acmart}

\AtBeginDocument{%
  \providecommand\BibTeX{{%
    \normalfont B\kern-0.5em{\scshape i\kern-0.25em b}\kern-0.8em\TeX}}}

\setcopyright{none}
\acmConference[RecSys in HR 2021]{Workshop on Recommender Systems for Human Resources at the ACM RecSys Conference on Recommender Systems (RecSys 2021)}{October 1, 2021}{Amsterdam, the Netherlands}
\acmDOI{}
\acmISBN{}
\copyrightyear{2021. Copyright for the individual papers remains with the authors. Copying permitted for private and academic purposes. This volume is published and copyrighted by its editors}
\acmPrice{}
\begin{document}

\title{conSultantBERT: Fine-tuned Siamese Sentence-BERT for Matching Jobs and Job Seekers} 

\author{Dor Lavi}
\email{dor.lavi@randstadgroep.nl}
\affiliation{%
  \institution{Randstad Groep Nederland}
  \city{Diemen}
  \country{The Netherlands}
}

\author{Volodymyr Medentsiy}
\email{volodymyr.medentsiy@randstadgroep.nl}
\affiliation{%
  \institution{Randstad Groep Nederland}
  \city{Diemen}
  \country{The Netherlands}
}

\author{David Graus}
\email{david.graus@randstadgroep.nl}
\affiliation{%
  \institution{Randstad Groep Nederland}
  \city{Diemen}
  \country{The Netherlands}
}

\renewcommand{\shortauthors}{Lavi et al.}

\begin{abstract}
  In this paper we focus on constructing useful embeddings of textual information in vacancies and resumes, which we aim to incorporate as features into job to job seeker matching models alongside other features. 
  We explain our task where noisy data from parsed resumes, heterogeneous nature of the different sources of data, and crosslinguality and multilinguality present domain-specific challenges. 
  
  We address these challenges by fine-tuning a Siamese Sentence-BERT (SBERT) model, which we call \texttt{conSultantBERT}, using a large-scale, real-world, and high quality dataset of over 270,000 resume-vacancy pairs labeled by our staffing consultants. 
  We show how our fine-tuned model significantly outperforms unsupervised and supervised baselines that rely on TF-IDF-weighted feature vectors and BERT embeddings. 
  In addition, we find our model successfully matches cross-lingual and multilingual textual content.
\end{abstract}




\maketitle

\section{Introduction}
Randstad is the global leader in the HR services industry. 
We support people and organizations in realizing their true potential by combining the power of today’s technology with our passion for people. 
In 2020, we helped more than two million job seekers find a meaningful job with our 236,100 clients. 
Randstad is active in 38 markets around the world and has top-three positions in almost half of these. 
In 2020, Randstad had on average 34,680 corporate employees and generated revenue of € 20.7 billion.

Each day, at Randstad, we employ industry-scale recommender systems to recommend thousands of job seekers to our clients, and the other way around; vacancies to job seekers. 
Our job recommender system is based on a heterogeneous collection of input data: curriculum vitaes (resumes) of job seekers, vacancy texts (job descriptions), and structured data (e.g., the location of a job seeker or vacancy). 
The goal of our system is to recommend the best job seekers to each open vacancy. 
In this paper we explore methods for constructing useful embeddings of textual information in vacancies and resumes. 

The main requirements of the model are (i) to be able to operate on multiple languages at the same time, and (ii) to be used to efficiently compare a vacancy with a large dataset of available resumes. 

Our end-goal is to incorporate these embeddings, or features derived from them, in a larger recommender system that combines a heterogeneous feature set, spanning, e.g., categorical features and real-valued features. 

\subsection{Problem setting}
\label{sec:problem-setting}
Several challenges arise when matching jobs to job seekers through textual resume and vacancy data.

First, the data we work with is inherently noisy.
On the one hand, resumes are user-generated data, usually (but not always) in PDF format. 
It goes without saying that parsing those files to plain text can be a challenge in itself and therefore out of scope for this paper. 
On the other hand, vacancies are usually structured formatted text. 

Second, the nature of the data differs. 
Most NLP research in text similarity is based on the assumption that two pieces of information are the same but written differently~\cite{10.1145/3440755}. 
However, in our case the two documents do not express the same information, but complement each other like pieces of a puzzle.
Our goal is to match two complementary pieces of textual information, that may not exhibit direct overlap/similarity. 

Third, as a multinational corporate that operates all across the globe, developing separate models for each market and language does not scale. 
Therefore, a desired property of our system is multilinguality; a system that will support as many languages as possible. 
In addition, as it is common to have English resumes in non-English countries (e.g., in the Dutch market around 10\% of resumes are in English), cross-linguality is another desired property, e.g., being able to match English resumes to Dutch vacancies. 

This paper is structured as follows.
First, in Section~\ref{sec:related-work} we summarize related work on the use of neural embeddings of textual information in the recruimtent/HR domain.
Next, in Section~\ref{sec:dataset-creation} we describe how we leverage our internal history of job seeker placements to create a labeled resume-vacancy pairs dataset. 
Then, in Section~\ref{sec:architecture} we describe how we fine-tune a multilingual BERT with bi encoder structure~\cite{reimers-gurevych-2019-sentence} over this dataset, by adding a cosine similarity log loss layer. 
Finally, in Section~\ref{sec:experiments} we describe how using the mentioned architecture helps us overcome most of the challenges described above, and how it enables us to build a maintainable and scalable pipeline to match resumes and vacancies. 
\section{Related Work}
\label{sec:related-work}
Neural embeddings are widely used for content-based retrieval, and embedding models became an essential component in the modern recommender system pipelines \cite{recsys1} \cite{Hassan2019BERTEU}. So the the main focus of our work is to construct embeddings of textual information in vacancy and resume, which could be then incorporated into another job-job seeker matching model and used along with other features, e.g. location and other categorical features. So we will focus on reviewing methods of embedding vacancies and resumes. The task of embedding resume and vacancy could be posed as creating domain-specific document embeddings. Although context-aware embeddings proved to outperform bag-of-words approaches in most of the NLP tasks in academia, the latter is still widely used in the industry.

\citeauthor{bian2020learningtomatch} propose to construct two sub-models with a co-teaching mechanism to combine predictions of those models. The first sub-model encodes relational information of resume and vacancy, and the second sub-model, which is related to our work, encodes textual information in resume and vacancy. 
Documents are processed per sentence, with every sentence being encoded using the CLS token of the BERT model. The Hierarchical Transformer is applied on top of sentence embeddings to get the document embeddings. The final match prediction is obtained by applying a fully-connected layer with sigmoid activation on concatenated embeddings of resume and vacancy.
\citeauthor{bhatia2019endtoend} propose to fine-tune BERT on the sequence pair classification task to predict whether two job experiences belong to one person or not. The proposed method does not require a dataset of labeled resume-vacancy pairs. The fine-tuned model is used to embed both the job description of the vacancy and the current job experiences of the job seeker. 
\citeauthor{zhao2021embeddingbasedrecommender} process words in resumes and vacancies using word2vec embeddings and domain-specific vocabulary. Word embeddings are fitted into a stack of convolutional blocks of different kernel sizes, on top of which \citeauthor{zhao2021embeddingbasedrecommender} apply attention to get the context vector and project that into the embedding space using the FC layer. The model is trained using the binary cross-entropy loss on the task of predicting a match between job seeker and vacancy.
\citeauthor{Ramanath2018TowardsDA} use supervised and unsupervised embeddings for their ranking model to recommend candidates to recruiter queries. The unsupervised method does not use the unstructured textual data but relies on the data stored in Linkedin Economic Graph \cite{LGE}, which represents entities such as skills, educational institutions, employers, employees, and relations among them. Candidates and queries are embedded using the graph neural models in this method.  
The supervised method embeds textual information in recruiters' queries and candidates' profiles using the DSSM \cite{DSSM} model. The DSSM operates on the character trigrams and is composed of two separate models to embed queries and candidates. The DSSM model is trained on the historical data of recruiters' interaction with candidates.

\citeauthor{zhu2018personjobfit} utilize skip-gram embeddings of various dimensionalities (64 for resume and 256 for a vacancy) to encode words, which are then passed through two convolutional layers. A pooling operation (max pooling for resume and mean pooling for vacancy) is applied to the output of convolutional layers to get the embeddings of resume and vacancy. The model is optimized using the cosine similarity loss. 
\citeauthor{Qin2018EnhancingPF} divide job postings into sections of ability requirements, and resumes into sections of experiences. The words are encoded using pre-trained embeddings and processed with bi-LSTM. To get the embeddings of job postings and resumes, they propose a two-step hierarchical pipeline. Every section is encoded using the attention mechanism, and finally to get embeddings of the job postings and resumes they run bi-LSTM on top of section encodings and aggregate bi-LSTM outputs using the attention mechanism. Additionally, during processing information in resumes \citeauthor{Qin2018EnhancingPF} propose to add encodings of job postings to emphasize skills in a resume relevant for a specific job. 

We can not directly compare our work with other approaches, because of the different datasets used. For example, \citeauthor{bhatia2019endtoend} and \citeauthor{Qin2018EnhancingPF} assume well-structured resumes, which is not the case in our situation.  \citeauthor{Ramanath2018TowardsDA} builds an embedding model for recruiters’ queries which are shorter than vacancies and lack the context provided in the vacancy. \citeauthor{bhatia2019endtoend} propose solution when limited amount of data is available. We on the other hand work in a setting of abundant data of heterogeneous nature, but at the same time resumes lack consistent structure, while vacancies are given in a more structured way. 
Additional issues that are not considered by most of the reviewed works are (i) \emph{cross-linguality}, so that we aim at predicting English resumes to Dutch vacancies if there is a potential match, and (ii) \emph{multilinguality}, where we aim to serve a single model for multiple languages. 
We address both issues with our approach. 
Next, \citeauthor{Qin2018EnhancingPF} and \citeauthor{zhu2018personjobfit} observe that an embedding model may benefit from constructing parallel pipelines to process resume and vacancy. 
Our approach relies on a shared embedding model of resumes and vacancies. 

While there are many organizations capable to train off the shelf transformers, not many of them have the availability of an abundance of high-quality labeled data. 
As a global market leader, we are situated in a unique position. 
We have both rich histories of high-quality interactions between consultants, candidates, and vacancies, in addition to having the content to represent those candidates and vacancies.
\section{Method}
Here we describe our method, more specifically, in Section~\ref{sec:dataset-creation} we describe how we acquire our labeled dataset of resume/vacancy-pairs. 
Next, in Section~\ref{sec:architecture} we describe our multilingual SBERT with bi-encoder and cosine similarity score as output layer. 

\subsection{Dataset creation}
\label{sec:dataset-creation}

We have rich history of interaction between consultants (recruiters) and job seekers (candidates). 
We define a positive signal any point of contact between a job seeker and consultant (e.g., a phone call, interview, job offer, etc.). 
Negative signals are defined by job seekers who submit their profile, but get rejected by a consultant without any interaction (i.e., consultant looks at the job seeker's profile, and rejects). 
In addition, since we have unbalanced dataset, for each vacancy we add random negative samples, which we draw randomly from our job seeker pool. 
This is done in spirit with other works, which also complement historical data with random negative pairs \cite{bian2020learningtomatch, zhao2021embeddingbasedrecommender, zhu2018personjobfit}. 

\begin{figure}
    \centering
    \includegraphics[width=\linewidth]{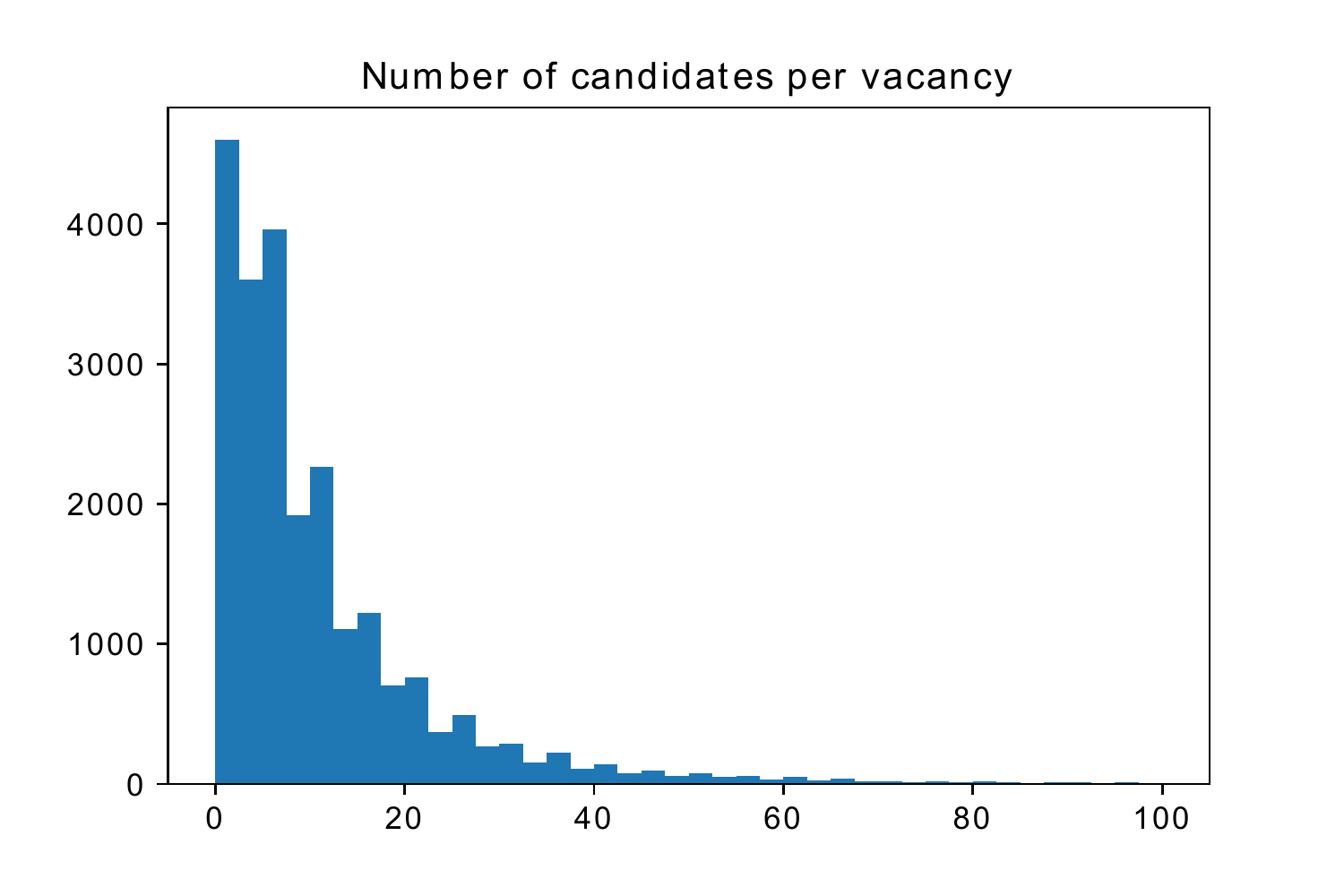}
    \caption{Histogram of the number of resumes per vacancy.}
    \label{fig:hist_cv_vac_plots}
\end{figure}

Our dataset consists of 274,407 resume-vacancy pairs, out of which 126,679 are positive samples, 109,724 are negative samples as defined by actual recruiters, and 38,004 are randomly drawn negative samples. We have 156,256 of unique resumes and 23,080 unique vacancy texts, which implies that one vacancy can be paired with multiple resumes. 
Figure~\ref{fig:hist_cv_vac_plots} shows the histogram of the number of resume-vacancy samples per vacancy. 
We see that for the majority of vacancies, we have a small number of paired resumes with approximately 10.5\% of our vacancies being paired with a single resume, and approximately 30\% of our vacancies being paired with at most three resumes. 

resumes are user-generated PDF documents which we parse with Apache Tika.\footnote{\url{https://tika.apache.org/}}
Overall, these parsed resumes can be considered quite noisy input to our model; there is a wide variation in format and structure of resumes, where common challenges include the ordering of different textual blocks, the diversity of their content (e.g., spanning any type of information from personalia, education, work experience, to hobbies), and parsing of tables and bullet points. 
At the same time, vacancies are usually well structured and standardized documents. 
They consist of on average 2,100 tokens (while resumes are on average longer and comprised of 2,500 tokens), and are roughly structured according to the following sections: 
job title, 
job description, 
job requirements, including required skills, 
job benefits, including compensation and other aspects, and 
company description, which usually includes information about the industry of the job offered in the vacancy.

\begin{figure*}[t]
    \centering
    \includegraphics[width=\linewidth,trim=-2.5cm 0 -2.5cm 0,clip]{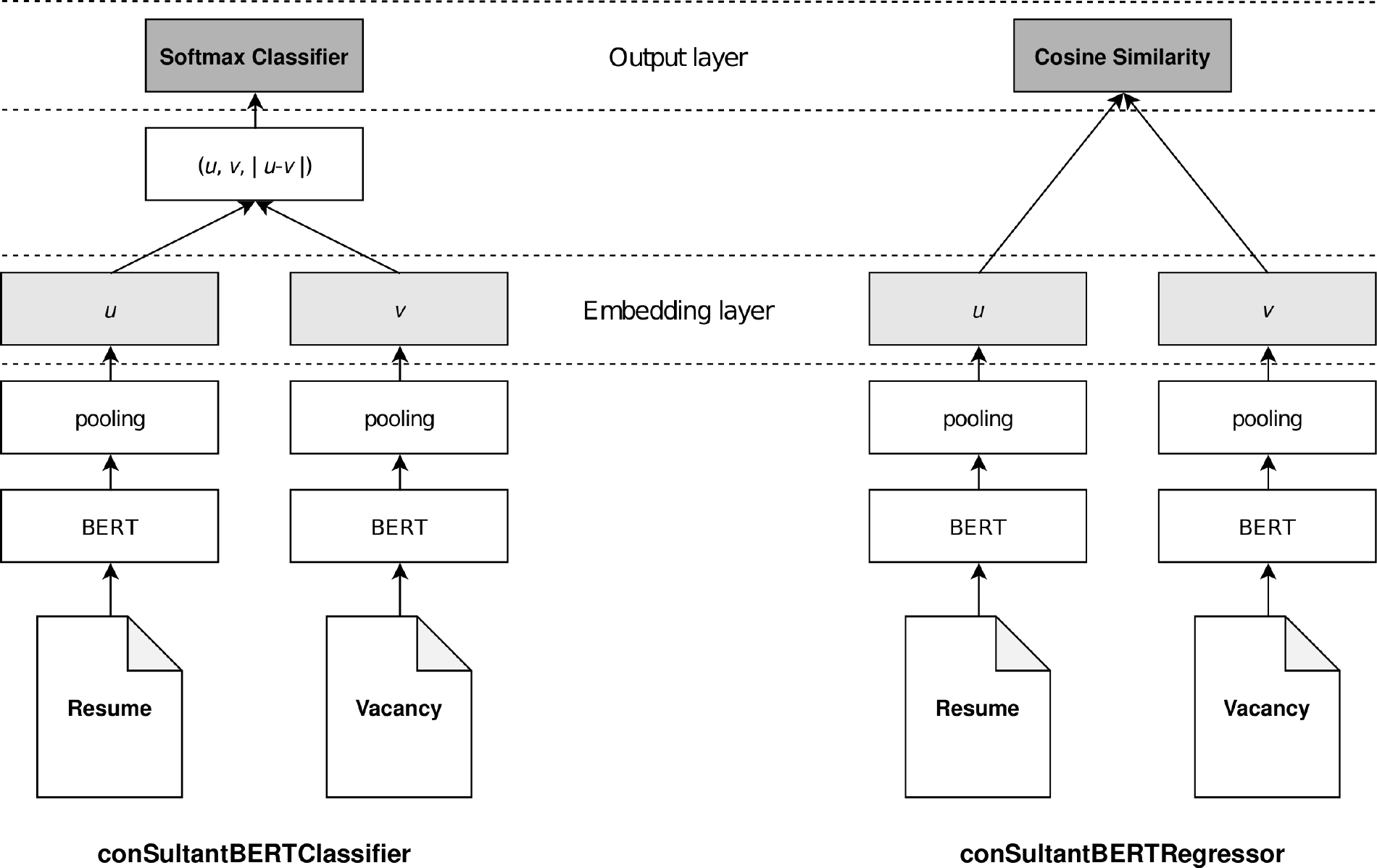}
    \caption{Our conSultantBERT architectures with classification objective (left) and regression objective (right), with resume text input on the left, and vacancy text input on the right-hand side. Image adopted from~\cite{reimers-gurevych-2019-sentence}. 
    Note that vacancy and resume differ at the level of words (vocabulary), but also format, structure, and semantics.}
    \label{fig:diagram}
\end{figure*}

\subsection{Architecture}
\label{sec:architecture}
Our method, dubbed \texttt{conSultantBERT} (as it is fine-tuned using labels provided by our consultants), utilizes the multilingual BERT~\cite{devlin2018bert} model pre-trained on Wikipedia pages of 100 languages.\footnote{We used the \texttt{bert-base-multilingual-cased} model from the HuggingFace library~\cite{wolf2020huggingfaces}.} 
We fine-tune it using the Siamese networks, as proposed by \citeauthor{reimers-gurevych-2019-sentence}. 
This method of fine-tuning BERT employs the bi-encoder structure, which is effective in matching vacancy with a large pool of available resumes. 

The original formulation of the Sentence-BERT (SBERT) model takes a pair of sentences as input, with words independently encoded in every sentence using BERT, which are aggregated by pooling to get the sentence embeddings.
After that, it either optimizes the regression loss, which is MSE loss between the cosine similarity score and true similarity label, or optimizes the classification loss, namely the cross-entropy loss.

Whereas SBERT is aimed at computing pairwise semantic similarity, we show that it can be applied effectively for our task of matching two heterogeneous types of data. Our fine-tuning pipeline is illustrated at the Figure \ref{fig:diagram}. So we pass resume and vacancy pairs to the siamese SBERT and experimented with classification and regression objectives.

\subsubsection{Document representation} 
Most transformer models, including SBERT, aim to model sentences as input, while we are interested in modeling  documents.

We experimented with several methods for document representation by using the embeddings of the pre-trained BERT model. First, we attempted to split our input documents into sentences to encode each sentence; 
we tried different sentence representations, first we used the pooled layer output from BERT to represent each sentence, which we then average to represent the document.
Here, we experimented with both simply averaging sentences, and weighted averaging (by sentence length). 

Next, we tried to take the mean of the last 4 layers of the \texttt{<CLS>} token, which is a special token placed at the start of each sentence, and considered a suitable representation for the sentence (according to~\citeauthor{devlin2018bert}). 
With these sentence representations, too, we represented the underlying document through averaging, both weighted and simple averaging.

Our final approach however, was more simple. 
We ended up treating the first 512 tokens of each input document as input for the SBERT model, ignoring sentence boundaries. 
To avoid trimming too much content, we pre-processed and cleaned our input documents by, e.g., removing non-textual content that resulted from parsing errors. 

\subsubsection{Fine-tuning method}\label{sec:fine-tuning}
We fine-tune our model on our 80\% training data split for 5 epochs, using a batch size of 4 pairs, and mean pooling, which was found during hyperparameter tuning on our validation set as optimal parameters.
To fine-tune the pre-trained BERT model, we need to trim the content of every resume and vacancy to 512 tokens, as the base model limits the maximum number of tokens to 512. 

\section{Experimental Setup}
\label{sec:experiments}
In this section, we describe the final datasets we use for training, validation, and testing in Section~\ref{sec:dataset}, the baselines and why we employ them in Section~\ref{sec:baselines}, our proposed fine-tuned \texttt{conSultantBERT} approach in Section~\ref{sec:consultantbert}, and finally, in Section~\ref{sec:evaluation-metrics} we explain our evaluation metrics and statistical testing methodology. 

\subsection{Dataset}\label{sec:dataset}
As described in Section~\ref{sec:dataset-creation}, our dataset consists of 274,407 resume-vacancy pairs. We split this dataset into 80\% train (219,525 samples), 10\% validation and 10\% test (27,441 samples each). 

We use our training set to (i) fine-tune the embedding models and train the supervised random forest classifiers, the validation set for the hyperparameter search of the SBERT hyperparams described in Section~\ref{sec:fine-tuning}, and report on performance on our test set. 

\subsection{Baselines}\label{sec:baselines}
As this paper revolves around constructing useful feature representations (i.e., embeddings) of the textual information in vacancy and resumes, we compare several approaches of generating these embeddings. 

\subsubsection{Unsupervised}
Our first baselines rely on unsupervised feature representations.
More specifically, we represent both our vacancies and resumes as either 
(i) TF-IDF weighted vectors (\texttt{TFIDF}), or
(ii) pre-trained BERT embeddings (\texttt{BERT}).

We then compute cosine similarities between pairs of resumes and vacancies, and consider the cosine similarity as the predicted ``matching score.''
Our TF-IDF vectors have 768 dimensions, which is equal to the dimensionality of BERT embeddings.\footnote{We experimented with increasing the dimensions of TF-IDF vectors, but it did not gave any substantial increase in performance (increasing dimensions 3-fold improved ROC-AUC by +0.7\%).} 
We fitted our TF-IDF weights on the training set, comprising both vacancy and resume data.  
As described in Section~\ref{sec:architecture}, we rely on BERT models pre-trained on Wikipedia from the HuggingFace library~\cite{wolf2020huggingfaces}. 

These unsupervised baselines help us to assess the extent to which the vocabulary gap is problematic, i.e., if vacancies and resumes use completely different words, both bag of words-based approaches such as TF-IDF weighting and embedding representations of these different words will likely show low similarity. 
Formulated differently, if word overlap or proximity between vacancies and resumes is meaningful, these baselines would be able to perform decently.

\subsubsection{Supervised}
Next, we present our two supervised baselines, where we employ a random forest classifier that is trained on top of the feature representation described above (\texttt{TFIDF+RF}, \texttt{BERT+RF}). 
These supervised baselines are trained on our 80\% train split, using the default parameters given by the scikit-learn library~\cite{scikit-learn}. 

We add these supervised methods and compare them to the previously described unsupervised baselines to further establish the extent of the aforementioned vocabulary gap, and the extent in which the heterogeneous nature of the two types of documents mentioned in Section~\ref{sec:problem-setting} plays a role. 
That is to say, if there is a direct mapping that can be learned from words in one source of data (vacancy or resume), to the other source, supervised baselines should be able to pick this up and outperform the unsupervised baselines.

\begin{table*}[t!]
  \begin{tabular}{lllccc}
    \toprule[1.5pt]
    \textbf{\#} & \textbf{Model} & \textbf{ROC-AUC} & \textbf{precision} & \textbf{recall} & \textbf{f1} \\
    \midrule[1.5pt]
    1. & \texttt{BERT+Cosine} &     0.5325 &     0.5057 &     0.5006 &     0.3379 \\
    2. & \texttt{TFIDF+Cosine} &     0.5730$^{*}$ &     0.5890 &     0.5016 &     0.3564 \\
    \midrule[1pt]
    3. & \texttt{BERT+RF} &     0.6978 &     0.6421 &     0.6363 &     0.6360 \\
    4. & \texttt{TFIDF+RF} &     0.7174 &     0.6581 &     0.6526 &     0.6527 \\
    \midrule[1pt]
    5. & \texttt{conSultantBERTClassifier+Cosine} &     0.7474 &     0.7001 &     0.6426 &     0.5994 \\
    6. & \texttt{conSultantBERTClassifier+RF} &     0.8366$^{*}$ &     0.7642 &     0.7643 &     0.7643 \\
    \midrule[0.5pt]
    7. & \texttt{conSultantBERTRegressor+Cosine} &     \textbf{0.8459}$^{*}$ &     0.7714 &     0.7664 &     0.7677 \\
    8. & \texttt{conSultantBERTRegressor+RF} &     0.8389 &     0.7684 &     0.7658 &     0.7666 \\
    \bottomrule[1.5pt]
  \end{tabular}
  \caption{Results of different runs. $^{*}$ denotes statistically significant difference from the alternative in the same group at $\alpha=0.01$. Best-performing run in \textbf{bold face}. }
  \label{tab:mother}
\end{table*}

\subsection{conSultantBERT \protect\footnote{The capitalized S in conSultant is a reference to SBERT.}}\label{sec:consultantbert}
Finally, we present our fine-tuned embedding model; \texttt{conSultant-}\\\texttt{BERT}, which we fine-tune using both the classification objective as the regression objective, as explained in Section~\ref{sec:architecture} and illustrated in Figure~\ref{fig:diagram}. 

Our intuition for using the classification objective is our task and dataset; which consists out of binary class labels, making the classification objective the most obvious choice.
At the same time, our work revolves around searching for meaningful embeddings, not necessarily solving the end-task in the best possible way, for which, we hypothesize, the model may benefit from the more fine-grained information that is present in the cosine similarity optimization metric. 

Similarly to our baselines, we consider (i) our fine-tuned model's direct output layer (i.e., the cosine similarity output layer) as ``matching score,'' between a candidate and a vacancy (\texttt{conSultantBERT-}\\\texttt{Regressor+Cosine}, \texttt{conSultantBERTClassifier+Cosine}), in addition to the predictions made by a supervised random forest classifier which we train on the embedding layer's output (see Figure ~\ref{fig:diagram}), yielding the following supervised models: \texttt{conSultantBERTReg-}\\\texttt{ressor+RF} and \texttt{conSultantBERTClassifier+RF}.
We explore the latter since we aim to incorporate our model in a production system alongside other models and features. 

\subsection{Evaluation}\label{sec:evaluation-metrics}
In order to compare our different methods and baselines, we turn to standard evaluation metrics for classification problems, namely we consider ROC-AUC as our main metric, as it is insensitive to thresholding and scale invariant. 
In addition, we turn to macro-averaged (since our dataset is pretty balanced) precision, recall, and F1 scores for a deeper understanding of the specific behaviors of the different methods. 

Finally, we perform independent student's $t$-tests for statistical significance testing, and set the $\alpha$-level at 0.01.
\section{Results}\label{sec:results}

See Table~\ref{tab:mother} for the results of our baselines and fine-tuned model. We first turn to our main evaluation metric ROC-AUC below, and next to the precision and recall scores in Section~\ref{sec:precision-recall}.

\subsection{Overall performance}\label{sec:results-overall}

\subsubsection{Baselines}\label{sec:baselines-performance}
As expected, using BERT embeddings or TF-IDF vectors in an unsupervised manner, i.e., by computing cosine similarity as a matching score, does not perform well. 
This confirms our observations about the challenging nature of our domain; word overlap or semantic proximity/similarity does not seem a sufficiently strong signal for identifying which resumes match a given vacancy. 

At row 3 and 4 we show the methods where a supervised classifier is trained using the input representations described above (\texttt{TFIDF+RF} and \texttt{BERT+RF}).
Here, we see that with a supervised classifier, both TF-IDF-weighted vectors as feature representation and pre-trained BERT embeddings vastly improve over the unsupervised baselines, with a +31.0\% improvement for BERT, and a +25.2\% improvement for the TF-IDF-weighted vectors. 
This suggests to some extent, a mapping can be learned between the two types of documents, as we hypothesized in Section~\ref{sec:baselines}.

\begin{figure*}[t]
    \centering
    \includegraphics[width=\textwidth]{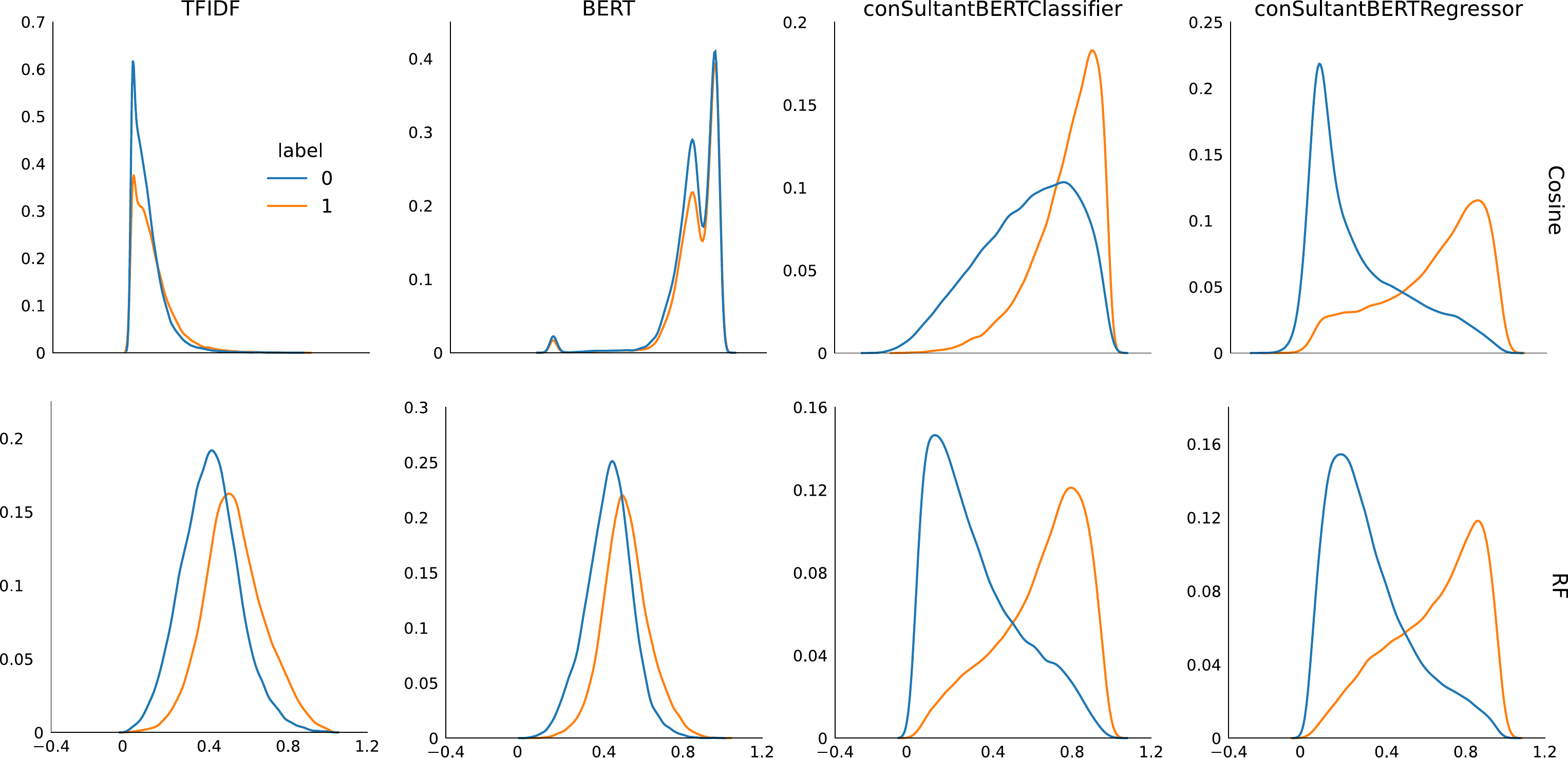}
    \caption{Density plots showing the distribution of Cosine similarity (denoted \texttt{Cosine}) scores (top row) and Random forest classifier (denoted \texttt{RF}) probability scores (bottom row) per label. Blue lines show the distributions for the negative class, and orange show distributions for the positive class.}
    \label{fig:density_plots}
\end{figure*}

\subsubsection{conSultantBERT}
Next, we turn to our consultantBERT runs in rows 5 through 8.
First, we consider models trained with the classification objective, in rows 5 and 6.
We note that fine-tuning BERT brings huge gains over the results the unsupervised pre-trained embeddings (\texttt{BERT+Cosine}), which is in line with previous findings~\cite{10.1007/978-3-030-32381-3_16}. 
Even the method that relies on direct cosine similarity computations on the embeddings learned with the classification objective (\texttt{conSultantBERTClassifier+Cosine}), outperforms our supervised random forest baselines in rows 3 and 4 with a 7.1\% and 4.2\% increase in ROC-AUC respectively. 
Adding a random forest classifier on top of those fine-tuned embeddings (\texttt{conSultantBERT}\\\texttt{Classifier+RF}) even further increases performance with a +11.9\% increase in ROC-AUC over our supervised baseline with pre-trained embeddings. 

As our primary goal is to get high quality embeddings for the down-stream application of generating useful feature representations for recommendation, alongside of which we can train whatever model we want with additional features types such as categorical, binary, or real-valued features, we are more interested in the former unsupervised, rather than the latter approach with random forest classifier. 

Finally, we turn to our fine-tuned \texttt{conSultantBERT} with the regression objective (\texttt{conSultantBERTRegressor}), to study the direct cosine similarity optimization objective. 
Looking at the last two rows, we find that conSultantBERT with regression objective performs similarly to conSultantBERT trained with the classification objective in supervised approach with a random forest classifier on top. 
In fact, the small difference turns out not to be statistically significant with a $p$-value of 0.2. 

What stands out most, is that the approach with cosine similarity outperforms all other runs, including significantly outperforming the \texttt{conSultantBERTClassifier+RF} approach. 
We explain this difference by the fact the architecture with the classifier objective having a learnable layer (the Softmax classifier in Figure~\ref{fig:diagram}). 
We drop this layer after the fine-tuning phase to yield our embeddings, losing information in the process. 
On the other hand, the architecture with the regression objective's had a non-learnable last layer, namely simply cosine similarity, which means all the necessary information has to propagate to the embedding layer. 

\subsection{Precision \& Recall}\label{sec:precision-recall}
When we zoom into the more fine-grained precision and recall scores, we observe the following: First, our unsupervised baselines in row 1 and 2 show how TFIDF-based cosine similarity yields substantially higher precision compared to cosine similarity with pre-trained BERT embeddings, keeping similar recall. 

Next, these same baselines with a supervised random forest classifier on top (row 3 and 4) show that in both cases the classifiers seem to perform somewhat similarly, irrespective of whether it is trained with TFIDF-weighted feature vectors or pre-trained BERT embeddings. 
With only a slight increase across precision and recall (around +2.5\%) for the TFIDF-based method, we conclude the feature space in itself does not provide substantially different signals for separating the classes. 

Then, comparing our fine-tuned models, rows 5 and 6 shows the SBERT model fine-tuned with the classification objective yields marginally higher precision compared to the random forest classifier trained on the SBERT's output layer. 
Row 7 and 8 show similar stable improvements across the board compared to the methods in rows 1--4.

Overall, we conclude that all of the supervised approaches (row 3 through 8) show roughly similar behavior with precision and recall balanced and substantial improvements over the unsupervised baselines. 
\begin{figure*}[t]
    \centering
    \includegraphics[width=\textwidth, trim=0 0 2.5cm 0, clip]{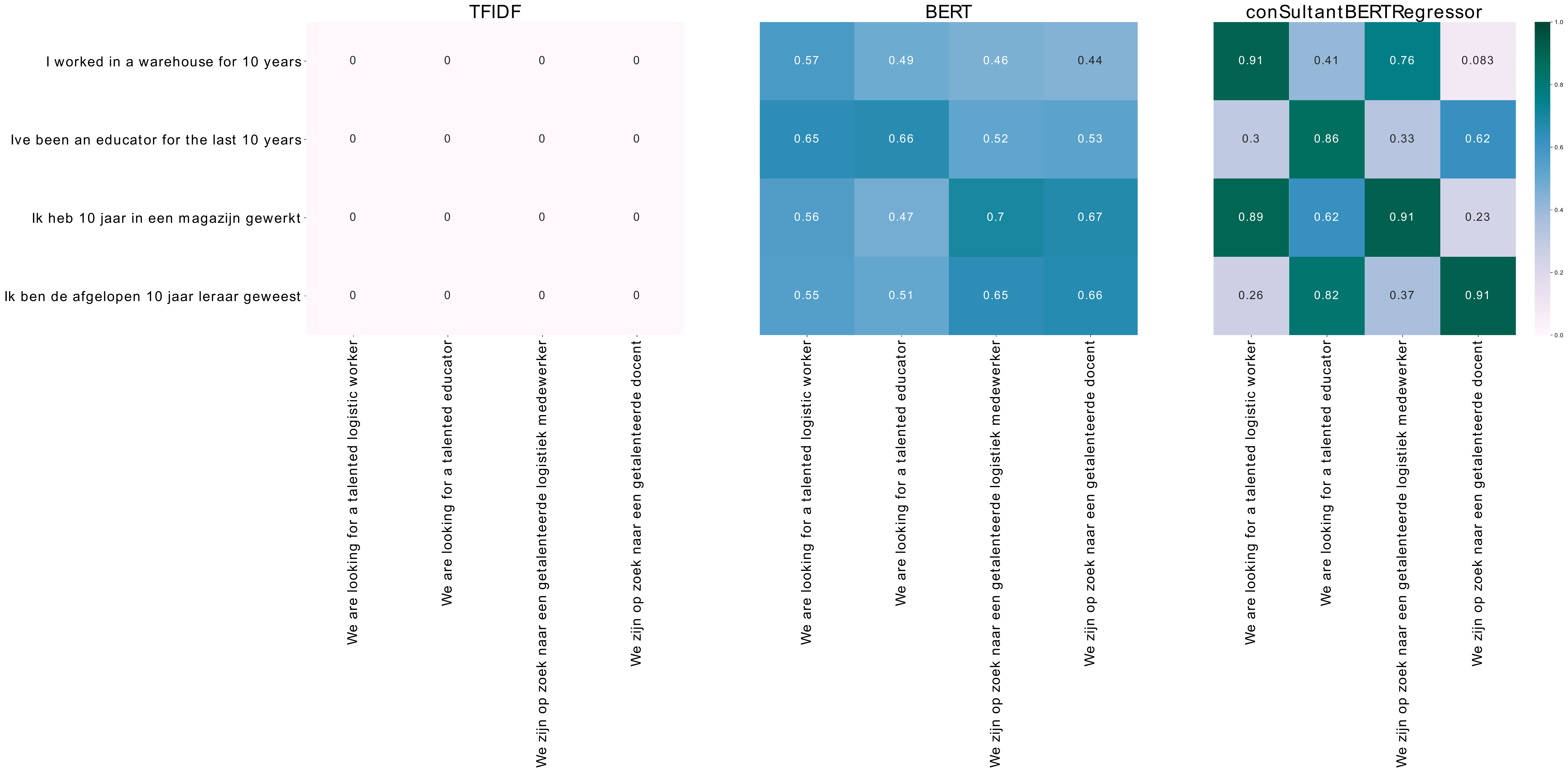}
    \caption{Heatmaps of cosine similarity between sentences from resumes and sentences from vacancies. The English sentences (first two rows, first two columns) and Dutch sentences (last two rows, last two columns) are each other's direct translations.}
    \label{fig:bert_sentences_cos_sim}
\end{figure*}

\section{Analysis and Discussion}
After analyzing the results in the previous section, in this section we take a closer look at the different methods, by studying the distributions of prediction scores across the two labels (positive and negative matches) in Section~\ref{sec:distributions}, and we zoom into another desired property of our embedding space: multilinguality (in Section~\ref{sec:multilinguality})

\subsection{Distributions}\label{sec:distributions}
In Figure~\ref{fig:density_plots} we plot the distributions of predicted scores per run, split out for positive and negative labels. 
The predicted scores correspond to cosine similarities in the case of \texttt{TFIDF}, \texttt{BERT}, \texttt{conSultantBERT*}, and probabilities of the random forest classifier in the case of \texttt{TFIDF+RF}, \texttt{BERT+RF}, and \texttt{conSultantBERT*+RF}.

A few things stand out, first, it is clear that fine-tuning the BERT embeddings in our conSultantBERT models yield a clearer separation in the predictions per class.
The left-most plots (TFIDF, both with Cosine similarities) show largely overlapping distributions.
The second column of plots (denoted BERT) shows largely similar patterns; with as main difference the unsupervised BERT having a bias towards higher scores compared to the unsupervised TFIDF. 

At the same time, while the distributions of scores for each label from the classifier with TF-IDF weighted vectors or pre-trained BERT embeddings do seem rather close, the model does succeed in separating the classes more often than not (as can be seen by the score improvements over the non-supervised baselines in Table~\ref{tab:mother}).

Observing the latter, though, makes clear that unsupervised similarity scores using default embeddings or TF-IDF weights does not allow a strong signal to separate both classes, which can be witnessed by the largely overlapping score distributions. 

We observed the difference in performance between our regression-optimized conSultantBERT with the classification-optimized one, in Section~\ref{sec:results-overall}. 
We now turn to comparing the prediction distributions of both models. 
We note how the classification-optimized conSultantBERT (top row, third plot from the left) seems to yield less separable cosine similarity scores compared to the regression-optimized one (top right plot). 
Compared to the three other conSultantBERT approaches, the former stands out. 
The random forest classifier trained on top of the embeddings, though, effectively learns again to separate between the classes, suggesting the embeddings keep distinguishing information.

\subsection{Multilinguality}\label{sec:multilinguality}
As a multinational corporate that operates all across the globe, developing a model per language is not scalable in our case. Therefore, a desired property of our system is \emph{multilinguality}; a system that will support as many languages as possible.

In some of the countries we operate, there is a high percentage of job seekers that are not native to that country. For example, many of the job descriptions in the Netherlands are in Dutch, however around 10\% of the resumes are in English. Due to that fact another expected property from our solution is to be \emph{cross-lingual}~\cite{10.1613/jair.1.11640}.

Classic text models, like TF-IDF and Word2vec, capture information within one language, but hardly connect between languages. Simply put, even if trained on multiple languages each language will have its own cluster in space. So \emph{``logistics''} in English and \emph{``logistiek''} in Dutch are embedded in a completely different point in space, even though the meaning is the same. 

Furthermore, we know that the language of resume correlates with nationality and therefore can be a proxy discriminator. Due to the impact of these systems and the risks of unintended algorithmic bias and discrimination, HR is marked as a high risk domain in the recently published EC Artificial Intelligence Act~\cite{european2021proposal}. To avoid discriminating against nationality we would like to recommend a candidate to the vacancy no matter which language the resume is written in. That is of course only if language is not a requirement for that vacancy.

In Figure~\ref{fig:bert_sentences_cos_sim} we see few examples for sentences that a candidate might write \emph{``I worked ...''} and few examples for vacancy \emph{``We are looking for ...''}. In order to demonstrate cross lingual and multilingual the same examples are written in both English and Dutch.

On the left hand side of Figure~\ref{fig:bert_sentences_cos_sim} (TFIDF) we can see that there is no match at all. This is due to the vocabulary gap, so by definition TFIDF can not match between \emph{``warehouse''} and \emph{``logistic worker''}. 

To solve the vocabulary gap we introduced BERT (center of Figure~\ref{fig:bert_sentences_cos_sim}), which we see does find similarity between candidate and vacancy sentences. 
However, it also hardly separates between the positive and negative pairs. 
Moreover, we can see a slight clustering around languages, so Dutch sentences have comparitively higher similarity to Dutch sentences, and likewise English sentences are more similar to each other.

On the right hand side of Figure~\ref{fig:bert_sentences_cos_sim} (\texttt{conSultantBERTRegressor}) we observe that the vocabulary gap is bridged, cross-lingual sentences are paired properly (e.g \emph{``I worked in a warehouse for 10 years''} and \emph{``We zijn op zoek naar een getalenteerde logistiek medewerker''} have high score), and finally, both Dutch to Dutch and English to English sentences are properly scored, too, thus achieving our desired property of multilinguality.
\section{Conclusion}\label{sec:conclusion}
In this work we experimented with various ways to construct embeddings of resume and vacancy texts.

We propose to fine-tune the BERT model using the Siamese SBERT framework on our large real-world dataset with high quality labels for resume-vacancy matches derived from our consultants' decisions. 
We show our model beat our unsupervised and supervised baselines based on TF-IDF features and pre-trained BERT embeddings. 
Furthermore, we show it can be applied for multilingual (e.g., English-to-English alongside Dutch-to-Dutch) and cross-lingual matching (e.g., English-to-Dutch and vice versa). 
Finally, we show that using a regression objective to optimize for cosine similarity yields more useful embeddings in our scenario, where we aim to apply the learned embeddings as feature representation in a broader job recommender system.

\begin{acks}
Special thanks to the rest of the SmartMatch team: Adam, Evelien, Najeeb, Sandra, Wilco, Wojciech, and Zeki. 
And Sepideh for her helpful comments. 
\end{acks}

\bibliographystyle{ACM-Reference-Format}
\bibliography{bibliography}


\end{document}